\documentclass{article}

\PassOptionsToPackage{numbers,compress}{natbib}
\usepackage[final]{caisc_2026}
\makeatletter
\renewcommand{\@noticestring}{}
\makeatother

\usepackage[utf8]{inputenc}
\usepackage[T1]{fontenc}
\usepackage{hyperref}
\usepackage{url}
\usepackage{booktabs}
\usepackage{amsfonts}
\usepackage{amsmath}
\usepackage{amssymb}
\usepackage{microtype}
\usepackage{xcolor}

\newtheorem{proposition}{Proposition}

\title{A Forced-Structure Reduction and Verifiable Bounds for
Conway's 99-Graph}

\author{%
  Aalok Thakkar \\
  Ashoka University\\
  \texttt{thakkar@ashoka.edu.in} \\
  \And
  Simone Severini \\
  UCL \\
  \texttt{s.severini@ucl.ac.uk} \\
}

\begin{document}

\maketitle

\begin{abstract}
Conway's 99-graph problem asks whether a strongly regular graph with parameters
$\mathrm{srg}(99,14,1,2)$ exists. We develop two complementary lines of
attack. Fixing one vertex, the conditions $\lambda=1$ and $\mu=2$ force its
neighbourhood to be a perfect matching and determine every edge between that
neighbourhood and the remaining vertices. For $(99,14,1,2)$, the unresolved
part is therefore a constrained $12$-regular graph on $84$ vertices. We encode
this reduction in CP-SAT and validate it by recovering the unique
$\mathrm{srg}(9,4,1,2)$. We also prove by exhaustive enumeration that no
circulant graph on $\mathbb{Z}/99$ satisfies more than $68.0\%$ of the CAISc
constraints, and we give a validated orbit formulation for prescribed
automorphisms. We then study the partial-score search problem. Fourteen human-designed search
configurations reached at most $69.43\%$. Separately, we supplied the scoring
function to an evolutionary program-search system. It produced a
degree-preserving $4$-vertex-switch tabu search whose best verified artifact
scores $70.73\%$. The generated move differs from those used in our own
searches and crosses a plateau that was stable under them. A tabu search
restricted to graphs with a prescribed symmetry group, first $\mathbb{Z}/33$
and then its subgroup of order $11$, reaches a $14$-regular graph scoring
$76.67\%$. These results do not resolve the existence problem, but they reduce the exact
search space and improve the best verified partial construction found in our
experiments.
\end{abstract}

\section{Introduction}
A \emph{strongly regular graph} $\mathrm{srg}(v,k,\lambda,\mu)$ is a
$k$-regular graph on $v$ vertices in which every adjacent pair has exactly
$\lambda$ common neighbours and every non-adjacent pair has exactly $\mu$.
The parameter set $(99,14,1,2)$ was recorded by Biggs in
1971~\cite{biggs1971}, and Conway later offered \$1000 for either a
construction or a proof of non-existence~\cite{conway2017,wiki99}. The problem
remains open. Its parameters satisfy the standard feasibility tests, including
the eigenvalue, Krein, and absolute-bound conditions~\cite{brouwertab,bcn1989},
while any solution would have to be non-vertex-transitive~\cite{cesarz2023}.

The CAISc 2026 Verifiable Problems track~\cite{caiscproblem} scores a submitted
$99\times99$ symmetric $0/1$ matrix by the fraction of graph constraints it
satisfies. There are $99$ degree constraints and one common-neighbour
constraint for each unordered pair, giving
\[
99+\binom{99}{2}=4950
\]
constraints in total. A score of $4950$ is exactly a solution to Conway's
problem.

We approach the problem in two ways.

First, we use the parameters themselves to reduce the search space. For any
$\mathrm{srg}(v,k,1,2)$, fixing one vertex forces its neighbourhood to be a
perfect matching. The outer vertices can then be identified with the
non-matched pairs in that neighbourhood, which determines all adjacency
between the inner and outer layers. For $(99,14,1,2)$, the only remaining
freedom is a constrained $12$-regular graph on $84$ vertices. We encode this
reduced problem in CP-SAT and validate the construction on
$\mathrm{srg}(9,4,1,2)$.

We also study restricted symmetry classes. An exhaustive search over circulant
graphs proves a maximum score of $68.0\%$. We then build an orbit model for
prescribed automorphisms, covering both fixed-point-free and
single-fixed-point actions. The model is validated on known strongly regular
graphs and applied to the single-fixed-point $\mathbb{Z}_7$ case, which
Behbahani and Lam had already excluded by computer~\cite{behbahani2011}.

Second, we study the partial-score problem directly. We tested fourteen
human-designed search configurations using local search, tabu search,
simulated annealing, evolutionary search, and CP-SAT neighbourhood
re-optimisation. Their best score was $69.43\%$.

We then ran a separate program-search experiment. The CAISc scoring function
was supplied as the fitness harness to an evolutionary program-search
system~\cite{alphaevolve}. It was not given the forced-structure reduction or
our hand-designed move operators. The generated program used a
degree-preserving $4$-vertex edge switch together with tabu memory and large
perturbations, and reached a verified score of $70.73\%$.

In version~3 we restrict the search to graphs with a prescribed symmetry
group, using Adrien Bouchet's tabu-search method (Section~\ref{sec:heuristic}). A tabu search over $\mathbb{Z}/33$
and then over its subgroup of order $11$ reaches a $14$-regular graph scoring
$76.67\%$.

Both improvements mean that the $69.43\%$ plateau came from the search
neighbourhoods we were using, not from a known bound on the problem. The exact
results and the heuristic results should therefore be read separately: the
circulant bound and structural reduction are proved, while the $70.73\%$ and
$76.67\%$ results are verified search artifacts.

The exact components use SAT, SMT, and CP-SAT
solvers~\cite{biere2021handbook,demoura2008z3,perron2024ortools}, which have
resolved cases of other open combinatorial
problems~\cite{heule2016pythagorean,konev2015erdos}. All search
artifacts are evaluated with the same scoring function.

\section{Exact structural results}

\subsection{Problem and scoring metric}
Let $A$ be the adjacency matrix and let $C=A^2$, so that $C_{ij}$ counts the
number of common neighbours of vertices $i$ and $j$. The strongly regular
conditions are equivalent to the matrix identity
\[
A^2+A-12I=2J,
\]
where $J$ is the all-ones matrix. The CAISc score therefore counts how many of
the $4950$ scalar instances of this identity are satisfied.

There is an immediate reason that low-order moment arguments do not settle the
problem. For any $14$-regular graph,
\[
\sum_{i<j}C_{ij}
=\sum_v\binom{\deg v}{2}
=99\binom{14}{2}
=9009.
\]
The target common-neighbour counts sum to
\[
693\cdot1+4158\cdot2=9009,
\]
exactly the same value. The feasible spectrum also fixes
$\sum_{i<j}C_{ij}^2=17325$, again matching the target. Thus neither the first
nor the second moment yields a contradiction. This also matters
computationally: once the graph is close to degree $14$, the $\lambda$- and
$\mu$-constraints are tightly coupled. One cannot improve $\mu$ cheaply in
isolation, because the total common-neighbour budget is already nearly
fixed. Among $14$-regular graphs, high partial scores therefore force the
graph to look increasingly close to strongly regular.

The metric itself does not enforce regularity, and $\mu$-constraints make up
$4158$ of the $4950$. The $10\times10$ rook's graph
$K_{10}\square K_{10}$ with one vertex deleted satisfies no degree constraint
and no $\lambda$-constraint, but every non-adjacent pair has $2$ common
neighbours except the $81$ pairs that lost one of theirs with the deleted
vertex. It scores
$3888/4950=78.5\%$, above every graph reported in this paper, without
resembling an $\mathrm{srg}(99,14,1,2)$. Scores are therefore comparable only
among $14$-regular graphs. Our $70.73\%$ artifact is $14$-regular; our
$69.43\%$ artifact satisfies $69$ of the $99$ degree constraints.

\subsection{An exhaustive bound for circulant graphs}
A circulant graph on $\mathbb{Z}/99$ is determined by a symmetric connection
set $S=-S$ with $|S|=14$, so it is automatically $14$-regular. By
vertex-transitivity, all pairs in the same difference class $\{d,-d\}$ behave
identically. The score is therefore
\[
99+99\cdot(\text{satisfied difference classes}),
\]
over the $49$ nonzero difference classes. A class $d$ is satisfied exactly
when either $d\in S$ and
\[
\lambda(d)=|S\cap(S+d)|=1,
\]
or $d\notin S$ and $\lambda(d)=2$. A perfect circulant would be a
$(99,14,1,2)$ partial difference set, which is known not to
exist~\cite{ma1994}. We therefore ask for the best partial one.

\begin{proposition}[Exhaustive]\label{prop:circ}
Over all $\binom{49}{7}=85{,}900{,}584$ symmetric connection sets, the maximum
number of satisfied difference classes is $33$. Hence the best circulant score
is
\[
99+99\cdot33=3366/4950=68.0\%.
\]
\end{proposition}

We verified Proposition~\ref{prop:circ} by complete enumeration using batched
FFT autocorrelation, which takes roughly $100$ seconds on a laptop. One
optimal connection set is
\[
S=\{\pm1,\pm2,\pm4,\pm15,\pm27,\pm36,\pm45\}.
\]
It satisfies all $99$ degree constraints, $198/693$ of the $\lambda$
constraints, and $3069/4158$ of the $\mu$ constraints. The algebra strongly
concentrates the non-edge common-neighbour counts around $2$: the $\mu$
success rate is $73.8\%$, compared with roughly $27\%$ for a random regular
graph. The non-cyclic abelian group of order $99$,
$\mathbb{Z}_3\times\mathbb{Z}_3\times\mathbb{Z}_{11}$, reaches the same
$33/49$ ceiling.

\subsection{The forced-structure reduction, validated on \texorpdfstring{$\mathrm{srg}(9,4,1,2)$}{srg(9,4,1,2)}}
\label{sec:reduction}
A large part of any $\mathrm{srg}(v,k,1,2)$ is forced once one vertex is fixed.
Let that vertex be $0$, with neighbourhood
\[
N(0)=\{1,\dots,k\}.
\]
Because $\lambda=1$, each neighbour of $0$ has exactly one common neighbour
with $0$. Therefore the subgraph induced by $N(0)$ is a perfect
matching~\cite{bcn1989}. Without loss of generality, fix this matching as
\[
(1,2),(3,4),\dots.
\]

Now consider a vertex outside $\{0\}\cup N(0)$. Since it is non-adjacent to
$0$ and $\mu=2$, it must have exactly two neighbours in $N(0)$. Moreover, the
$\lambda$ and $\mu$ constraints on pairs inside $N(0)$ force the outer
vertices to be in bijection with the non-matched pairs of neighbours. Thus an
outer vertex can be labelled by a pair $\{a,b\}$, and its adjacency to the
inner layer is completely determined: it is adjacent to exactly $a$ and $b$.

The only remaining freedom is the graph induced by the outer vertices. This is
a $(k-2)$-regular graph on
\[
M=\binom{k}{2}-k/2
\]
vertices, subject to constraints determined by the pair labels. Let
$m(2i)=2i+1$ denote the matching partner map, and let $P_u$ be the pair
label of outer vertex $u$. For two outer vertices $u,v$, let
$s=|P_u\cap P_v|\in\{0,1\}$. Then:
if $u\sim v$, the number of common outer neighbours must be $1-s$; if
$u\not\sim v$, it must be $2-s$. For an inner vertex $a$ and an outer vertex
$u$, if $a\in P_u$, exactly one outer neighbour of $u$ contains $a$; if
$a\notin P_u$, then
\[
[m(a)\in P_u]
+\#\{\text{outer neighbours of }u\text{ containing }a\}
=2.
\]

For $(99,14,1,2)$, the entire problem therefore reduces to a $12$-regular
graph on $84$ outer vertices.

We encode the reduced problem in CP-SAT and add lex-leader symmetry breaking
for the group $B_{k/2}$ that permutes the $k/2$ matched pairs and swaps the
two vertices within each pair. For $k=14$, this group has order
\[
2^7\cdot7!=645{,}120,
\]
and we use its $13$ natural generators. The full reduced model contains
$379{,}987$ Boolean variables and $761{,}221$ constraints.

We validated the complete pipeline on the smaller case
$\mathrm{srg}(9,4,1,2)$, where it recovers the unique graph in milliseconds,
both with and without symmetry breaking. On the $(99,14,1,2)$ instance, the
solver neither finds a graph nor proves infeasibility. That is what one
should expect for an open problem, so we treat the encoding as a validated
framework rather than as evidence for existence or non-existence.

\subsection{Prescribed automorphisms}
A natural next step is to ask whether the graph exists under a prescribed
automorphism. Let $G$ be the automorphism group of a putative
$\mathrm{srg}(99,14,1,2)$. Cesarz and Woldar~\cite{cesarz2023}, building on
the orbit-matrix method of Behbahani and Lam~\cite{behbahani2011}, prove that
if $2\mid|G|$, then $|G|\mid6$, and if $7\mid|G|$, then
$G\cong\mathbb{Z}_7$. These proofs are computer-free.
Earlier, Behbahani and Lam~\cite{behbahani2011} had shown by computer that
every automorphism of prime order has order $2$ or $3$, as recorded in the
published version of~\cite{cesarz2023}. In particular, no
$\mathrm{srg}(99,14,1,2)$ has an automorphism of order $7$. Together with the
exclusion of groups of order $6$ and $9$~\cite{crnkovic2020}, this leaves
$|G|\in\{1,2,3\}$. Version~2 of this paper described the $\mathbb{Z}_7$ case
as open; that was an error.

Because non-fixed vertices under an order-$7$ action lie in $7$-cycles, the
number of fixed points satisfies
\[
f\equiv99\equiv1\pmod7.
\]
The smallest possibility is therefore one fixed point together with fourteen
orbits of size $7$.

We built an orbit-based existence encoding for a prescribed automorphism of
prime order $p$. Label the non-fixed vertices $(i,a)$, where $i$ indexes the
$B$ orbits of size $p$ and $a\in\mathbb{Z}/p$ ($B=99/p$ for a fixed-point-free
action and $B=98/p$ with one fixed point). An invariant graph is
block-circulant: adjacency between $(i,a)$ and $(j,b)$ depends only on
$(i,j,b-a\bmod p)$. For each pair class, the strongly regular condition
collapses to
\[
\textit{common}+\textit{adjacency}=2,
\]
which simultaneously encodes $\lambda=1$ and $\mu=2$.

The framework supports both fixed-point-free actions and actions with a single
fixed point. In the latter case, the fixed vertex must be adjacent to complete
orbits; for $p=7$, the degree condition $14=2\cdot7$ forces it to join exactly
two such orbits. We validated the implementation by reconstructing and
re-verifying $\mathrm{srg}(9,4,1,2)$ under both a fixed-point-free
$\mathbb{Z}_3$ action and an order-$2$ action with one fixed point, and by
recovering the Paley graph $\mathrm{srg}(13,6,2,3)$ under an order-$3$ action
with one fixed point.

Since the single-fixed-point $\mathbb{Z}_7$ model is known to be infeasible,
it tests whether a general-purpose solver can reproduce a known exclusion. A
complete run would return an infeasibility certificate. CP-SAT instead
returned \textsc{unknown} even after a $48$-hour run on $14$ cores. The fixed-point-free $\mathbb{Z}_3$ model, with
$33$ orbits, likewise remained \textsc{unknown} within $1800$ seconds.

We regard this as a negative result about the method rather than about the
graph. Even on a subproblem that Behbahani and Lam settled with orbit-matrix
methods, the general-purpose CP-SAT model did not decide the instance. The fact that the outcome persisted across a $96\times$ larger
time budget suggests that the main limitation is structural rather than
insufficient runtime. This points toward more specialised
orbit-matrix and eigenvalue-interlacing machinery.

\section{Heuristic search and program search}
\label{sec:heuristic}
For general asymmetric graphs, we built an $O(\deg)$ incremental search engine
that maintains the adjacency matrix $A$, the common-neighbour matrix
$C=A^2$, the squared error
\[
\mathrm{SE}=\sum_{i<j}(t_{ij}-C_{ij})^2,
\]
where $t_{ij}=1$ for edges and $2$ for non-edges, and the exact-match score.
This squared-error objective was also used in an earlier evolutionary
approach~\cite{hutnyk2019}. With these quantities cached, a single edge
toggle can be updated in roughly $O(14)$ time.

The exact-match score by itself is extremely flat: many local moves leave it
unchanged, giving little guidance to local search. We therefore optimise a
blended objective
\[
O(A)=\text{real}(A)-\alpha\,\mathrm{SE}(A),
\]
where the exact score remains primary and the squared-error term supplies a
direction across plateaus. We combine this with an island-model evolutionary
algorithm and degree-preserving crossover.

Across fourteen human-designed configurations
(Table~\ref{tab:methods}), the best score was $69.43\%$. High-scoring solutions consistently sit near
$\lambda\approx53\%$ and $\mu\approx72\%$. The optimal circulant is a strict
$2$-opt local maximum, and large-neighbourhood CP-SAT re-optimisation over
$14$-vertex chunks produces only lateral moves. Starting from the
$69.43\%$ artifact, a noisier min-conflicts run accepted
$1.57\times10^6$ moves without gaining even one additional satisfied
constraint.

\paragraph{An evolutionary-code-search program reaches $70.73\%$.}
We therefore ran a separate program-search experiment. The CAISc scoring
function was supplied as the fitness harness to an evolutionary program-search
system~\cite{alphaevolve}; the system was not given our forced-structure
reduction or our hand-designed move operators. Searching over programs instead
of graphs is a form of program
synthesis~\cite{synt2020,thakkar2021,mobius2023}, here guided by a numeric
score instead of examples or a formal specification. The generated program produced a
different search strategy whose best verified artifact scores
\[
3501/4950=70.73\%.
\]
Every candidate returned by the program search was evaluated by the same
independent scorer used for the human-designed methods.

Its core move is a degree-preserving $4$-vertex edge switch. Given edges
$uv$ and $xy$, with $x\neq u,v$, $u\not\sim x$, and $v\not\sim y$, it replaces
$\{uv,xy\}$ by $\{ux,vy\}$. The search is guided by a penalty gradient derived
from adjacency and common-neighbour deviations, uses a tabu list of length
$2000$, and applies a large perturbation of $20$--$40$ random switches after
$150$ stagnant iterations.

This move differs qualitatively from the $2$-opt edge toggles used in our own
searches. It preserves $14$-regularity by construction and can jump directly
between configurations that are not adjacent under the simpler move set. In
practice, that was enough to escape plateaus on which our min-conflicts and
blended simulated-annealing runs had stalled.

\begin{table}[t]
  \caption{Representative results across the configurations we ran (rows
  group related methods). The fourteen human-designed methods of version~2 top
  out at $69.43\%$; the program-search-generated $4$-vertex switch tabu search
  reaches $70.73\%$; the orbit tabu search added in version~3 reaches
  $76.67\%$.}
  \label{tab:methods}
  \centering\small
  \begin{tabular}{lll}
    \toprule
    Method & character & best score \\
    \midrule
    Exhaustive circulant ($\mathbb{Z}/99$) & proven bound & 68.0\% \\
    Cayley search ($\mathbb{Z}_3{\times}\mathbb{Z}_3{\times}\mathbb{Z}_{11}$) & search & 68.0\% \\
    Block CP-SAT, $\mathbb{Z}/11,\mathbb{Z}/9,\mathbb{Z}/3$ & exact, symmetric & 68.0\% \\
    Full MaxSAT ($504{,}504$ vars) & exact & 68.0\% \\
    Degree-preserving 2-opt SA & heuristic (human) & $\approx56\%$ \\
    Tabu / ILS & heuristic (human) & 69.3\% \\
    Min-conflicts; blended SA $+$ island EA & heuristic (human) & 69.43\% \\
    $4$-vertex switch tabu (program-search) & heuristic (generated) & 70.73\% \\
    Orbit tabu, $\mathbb{Z}/33$ then order-$11$ subgroup & heuristic, symmetric & \textbf{76.67\%} \\
    \bottomrule
  \end{tabular}
\end{table}

Table~\ref{tab:methods} summarises the frontier. The $70.73\%$ artifact satisfies
all $99$ degree constraints, $549/693$ of the $\lambda$ constraints, and
$2853/4158$ of the $\mu$ constraints. The earlier $69.43\%$ plateau was a
strict local optimum under our own $2$-opt-based move sets, but it was not a
global bound: the generated heuristic crosses it by $64$ constraints.

This suggests that further progress near the $70\%$ level may depend at least
as much on the geometry of the move set as on the choice of search
meta-heuristic. The local-optimum landscape appears to have substantial
structure, and seemingly small changes in the neighbourhood relation can open
paths that are invisible to simpler local moves.

To our knowledge, no earlier work reports a partial-credit score under the
track's exact constraint metric. The evolutionary search of
Hutnyk~\cite{hutnyk2019}, for example, optimised squared error rather than the
number of satisfied constraints.

After version~2 of this paper appeared, Adrien Bouchet (personal
communication) found two $14$-regular graphs with higher scores, by tabu
search over the pair orbits of a prescribed group with every move scored
exactly. The first scores $3729/4950=75.33\%$ ($429/693$ $\lambda$,
$3201/4158$ $\mu$). Its vertices are $(i,a)$ with $i\in\{0,1,2\}$ and
$a\in\mathbb{Z}/33$, and $(i,a)\sim(j,a+d)$ for $d\in S_{ij}$, where
$S_{00}=\{\pm4,\pm5,\pm8,\pm11\}$, $S_{11}=S_{22}=\{\pm1,\pm7,\pm10,\pm13\}$,
$S_{01}=\{0,14,31\}$, $S_{02}=\{12,26,28\}$ and $S_{12}=\{26,30,31\}$. One
degree-preserving switch turns it into a graph with trivial automorphism group
that scores $3699/4950=74.73\%$ ($433/693$ $\lambda$, $3167/4158$ $\mu$). We
rebuilt the first graph from this description and recomputed both scores and
the trivial automorphism group independently. Imposing a symmetry group on the
search space did better than our unstructured searches, including the
generated $4$-vertex switch.

We then applied Bouchet's tabu-search method ourselves. Our search fixes a permutation group,
toggles one orbit of vertex pairs per move, scores every move exactly, and
penalises deviation from degree $14$. Over $\mathbb{Z}/33$ acting freely with
three orbits, as in the $3729$ graph ($147$ pair orbits) it reached $3696$. Restarting from such a graph in
the larger family invariant only under the subgroup of order $11$ ($441$ pair
orbits) raised the score to
\[
3795/4950=76.67\%,
\]
with all $99$ degree constraints, $330/693$ $\lambda$-constraints and
$3366/4158$ $\mu$-constraints satisfied. We confirmed the score with an
independent scorer. From the $3729$ graph, the same procedure found nothing
higher, and searches in the order-$11$ family from random starts reached only
about $3170$, so the gain depends on starting from a good $\mathbb{Z}/33$
graph. Like the $3729$ graph, this graph has an automorphism of order $11$, so
it cannot be an $\mathrm{srg}(99,14,1,2)$~\cite{behbahani2011}.

\paragraph{Agent trajectory.}
The coding agent first attempted a full SMT formulation, then moved to CP-SAT
after the initial model stalled on global structural constraints. That change
made the circulant enumeration and block models practical and led to the
forced-structure and orbit formulations reported above. After checking the
literature on the open status and automorphism restrictions, we restricted the
paper's claims to independently verifiable results. The final exact-model
attempt focused on the single-fixed-point $\mathbb{Z}_7$ case and remained
\textsc{unknown} after $48$ hours. The evolutionary code-search experiment was
separate: it received only the scoring function as its fitness harness, and
the returned program was evaluated by the same verifier used for all other
artifacts. The version-3 orbit search was also written and run with an AI
coding assistant, and its best graph was rescored by a separate
implementation of the metric.

\section{Discussion and limitations}
The exact and heuristic results have different meanings.

The circulant bound is exhaustive, and the forced-structure reduction is exact.
The prescribed-automorphism formulation is also exact as an encoding, although
general-purpose CP-SAT does not decide the cases we tested. In particular,
the single-fixed-point $\mathbb{Z}_7$ model, which is known to be
infeasible~\cite{behbahani2011}, remained \textsc{unknown} after a $48$-hour
run on $14$ cores.

The $70.73\%$ and $76.67\%$ artifacts are search results, not bounds and not
evidence for or against existence. Both improve on a stable $69.43\%$ plateau
reached by our human-designed searches. The generated $4$-vertex switch gives
the search access to configurations that our earlier move sets did not reach
directly, and the prescribed symmetry reduces the search to a few hundred
binary choices. The $76.67\%$ graph has an automorphism of order $11$, which no
$\mathrm{srg}(99,14,1,2)$ can have.

A perfect score of $4950$ would construct the graph. Any provable global upper
bound below $4950$ would prove non-existence. For that reason, sufficiently
strong exact progress on the scoring problem eventually becomes the original
open problem.

\section{Conclusion}
We give several verifiable results on Conway's 99-graph. The main structural result reduces the general
$\mathrm{srg}(99,14,1,2)$ problem to a constrained $12$-regular graph on
$84$ vertices. We also prove an exhaustive $68.0\%$ upper bound for circulant graphs and give a validated orbit formulation for prescribed automorphisms. On the search side, fourteen human-designed configurations reached at most
$69.43\%$. An evolutionary program-search system produced a different
degree-preserving $4$-vertex-switch tabu search and reached a verified
$70.73\%$. Restricting the search to graphs with a prescribed symmetry group
raised the best $14$-regular score to $76.67\%$. None of these results
settles Conway's problem. They do, however, reduce the
exact search space, and the partial-score frontier moved twice when the search
neighbourhood changed.

\begin{ack}
AT was supported by the Anusandhan National Research Foundation (ANRF),
Government of India, under the Prime Minister Early Career Research Grant
ANRF/ECRG/2025/001136/ENS. We thank the AlphaEvolve
team~\cite{alphaevolve} at Google DeepMind for access to the
evolutionary code-search system used to generate the $70.73\%$ heuristic.
We thank Adrien Bouchet for pointing out the exclusion of order-$7$
automorphisms in~\cite{behbahani2011}, the rook's-graph example, and the
graphs scoring $3729$ and $3699$, and for the orbit tabu-search method that
produced our $3795$ graph.
Preliminary results on this problem were presented at the 1st Conference
For AI Scientists (CAISc 2026).
\end{ack}

\small
\sloppy
\setlength{\emergencystretch}{3em}
\hbadness=10000

\end{document}